\documentclass{article}

\usepackage{amsmath}
\usepackage{PRIMEarxiv}
\usepackage{enumitem}
\usepackage{multirow}
\usepackage{array}

\usepackage[utf8]{inputenc} 
\usepackage[T1]{fontenc}    
\usepackage{hyperref}       
\usepackage{url}            
\usepackage{booktabs}       
\usepackage{amsfonts}       
\usepackage{nicefrac}       
\usepackage{microtype}      
\usepackage{lipsum}
\usepackage{fancyhdr}       
\usepackage{graphicx}       
\graphicspath{{media/}}     

\usepackage{amssymb}
\usepackage{float} 
\usepackage{breqn}
\usepackage{amsthm}

\title{Diversity Enhances an LLM's Performance in RAG and Long-context Task}

\author{
  \textbf{Zhichao Wang*}, \textbf{Bin Bi}, \textbf{Yanqi Luo}, \textbf{Sitaram Asur}, \textbf{Claire Na Cheng}\\
  Salesforce \\
  zhichaowang@salesforce.com
}

\begin{document}
\maketitle
\def\thefootnote{}\footnotetext{$*$: corresponding author\\
\hspace*{5.5mm}github code: \url{https://github.com/ZhichaoWang970201/DIVERSITY-ENHANCES-LLM}}

\begin{abstract}
The rapid advancements in large language models (LLMs) have highlighted the challenge of context window limitations, primarily due to the quadratic time complexity of the self-attention mechanism (\(O(N^2)\), where \(N\) denotes the context window length). This constraint impacts tasks such as retrieval-augmented generation (RAG) in question answering (Q\&A) and long context summarization. A common approach involves selecting content with the highest similarity to the query; however, this often leads to redundancy and the exclusion of diverse yet relevant information. 
Building on principles from Maximal Marginal Relevance (MMR) and Farthest Point Sampling (FPS), we integrate diversity into the content selection process. Our findings reveal that incorporating diversity substantially increases the recall of selecting relevant sentences or chunks before LLM-based Q\&A and summarization. These results highlight the importance of maintaining diversity in future LLM applications to further improve summarization and Q\&A outcomes.
\end{abstract}

\begin{figure}[t]
    \centering
    \includegraphics[width=0.95 \textwidth,height=\textheight,keepaspectratio]{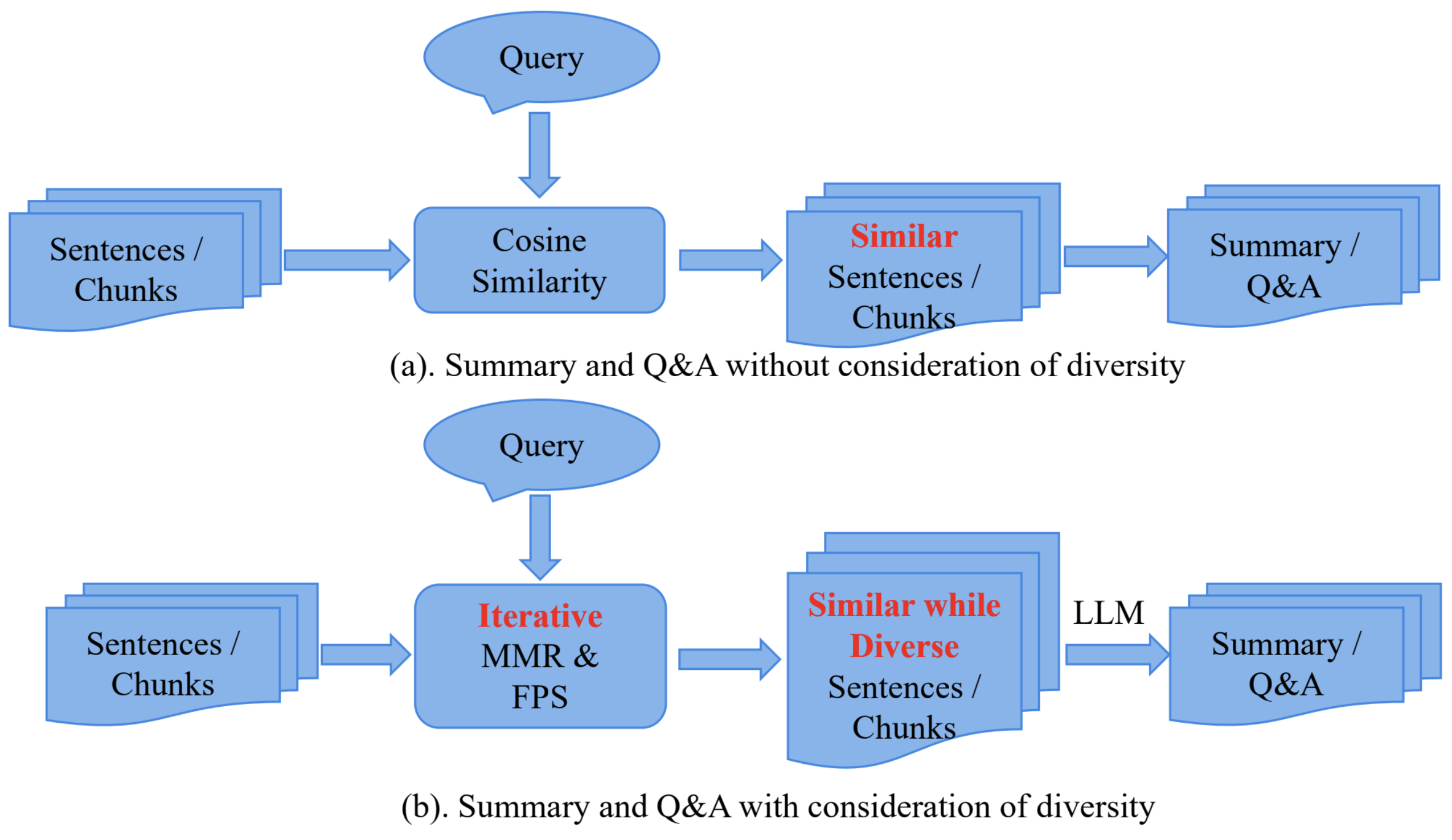}
    \caption{For both Q\&A and summarization tasks, the initial dataset is divided into sentences or chunks, and corresponding embeddings are extracted. In a traditional pipeline, query embeddings are generated and used to select relevant materials to LLMs for downstream tasks. In contrast, methods like MMR and FPS incorporate diversity in a greedy manner when selecting relevant sentences. This approach increases the recall of the correct answer within the chosen sentences or chunks.}
    \label{fig: placeholder}
\end{figure}

\section{Introduction}
The remarkable success of Transformer models \cite{vaswani2023attention}, BERT \cite{devlin2019bert}, and GPT \cite{openai2024gpt4} can be largely attributed to their robust self-attention mechanisms. However, the self-attention module's quadratic time complexity, \(O(N^{2})\), where \(N\) represents the context window length, has imposed limitations on the size of the context window.

Recent advances in LLMs have partially addressed this constraint. For instance, GPT-3.5 demonstrates the capability to process context windows of up to 16,385 tokens, while GPT-4 extends this capacity to an impressive 128,000 tokens. Despite these notable improvements, the challenge of processing even longer sequences remains a critical area of research for several compelling reasons. First, many real-world applications, such as question-answering systems operating on extensive datasets, cannot accommodate entire document collections within the LLM's context window. This limitation has led to the development of Retrieval-Augmented Generation (RAG) systems \cite{lewis2021retrievalaugmentedgenerationknowledgeintensivenlp}, which selectively retrieve and process relevant text segments for specific queries. Second, while current context window sizes may suffice for conventional Natural Language Processing (NLP) tasks, they prove inadequate for high-frequency signal processing applications. For example, audio processing and medical vibrational signal analysis often require handling data streams with sampling rates reaching one million samples per second, far exceeding current context window capabilities \cite{gu2024mambalineartimesequencemodeling}. Furthermore, empirical studies have revealed a concerning trend: LLM performance tends to degrade as input lengths approach the maximum context window capacity, highlighting the need for more robust solutions to long-sequence processing \cite{nvidia2024nemotron4340btechnicalreport}.

Various strategies have been devised to address the limited context window issue in LLMs. Longformer \cite{beltagy2020longformer} applies attention to immediate local neighbors, reducing the time complexity from \(O(N^{2})\) to \(O(NM)\), with \(M\) representing the considered neighbors. An alternative strategy is to expand the context window at inference \cite{jin2024llmmaybelonglmselfextend}. Lastly, a more straightforward approach is to first select the most relevant documents while ensuring they fit within the LLM's context window. For a given query, multiple documents are split into smaller chunks or sentences. The embeddings for both the query and the split documents are then computed. Similarity metrics, such as cosine similarity or Euclidean distance, are subsequently used to identify the most relevant sentences.

However, relying solely on the similarity between a query and segmented documents can result in overlooking critical information due to excessive focus on similar content. Previous studies have introduced greedy algorithms, such as MMR \cite{10.1145/290941.291025} and FPS \cite{qi2017pointnet}, to improve diversity during the selection process. In this paper, we aim to demonstrate the significance of diversity in long context summarization and RAG-based Q\&A tasks at multiple levels: sentence-level for single documents, chunk-level across entire datasets, and sentence-level in summarization.

The contributions of this paper are summarized as follows:

\begin{enumerate}[label=\arabic*.]
    \item We demonstrate the benefits of diversity using MMR and FPS with proper hyperparameters, i.e., $\alpha$ and $w$ on downstream tasks, including Q\&A and summarization.
    \item We discover that MMR achieves slightly better recall than FPS while maintaining significantly lower latency.
    \item We prove the ordering selected sentences within the original document and ordering selected chunks based on the scores has the best downstream performances.
\end{enumerate}

\section{Methodology}
In this section, we will start with a brief introduction of MMR and FPS to consider diversity during the search process. Then, the integration with LLM will be discussed.

\subsection{FPS and MMR for Diversity}

\begin{figure}[t]
    \centering
    \includegraphics[width=0.95\textwidth,height=\textheight,keepaspectratio]{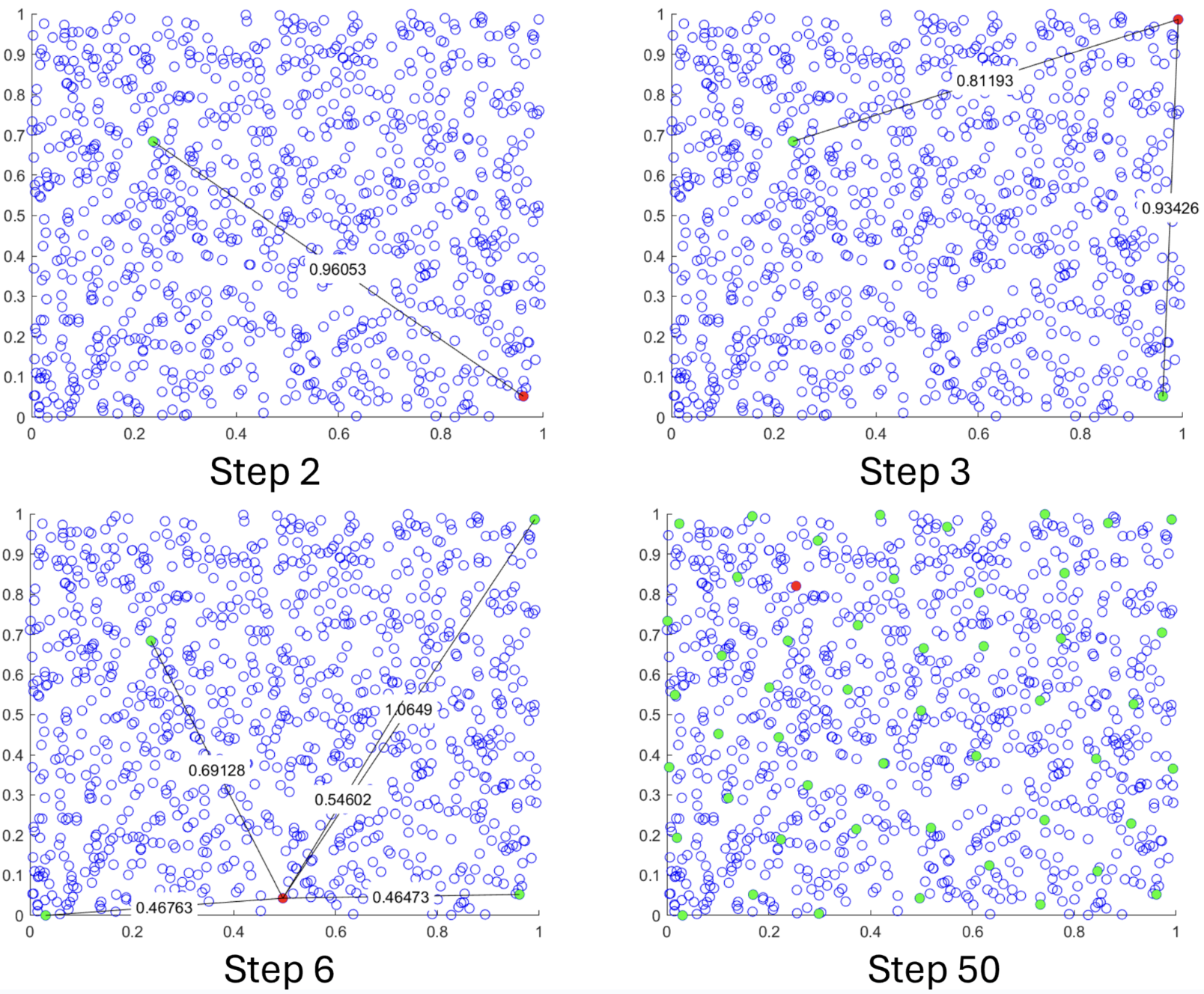}
    \caption{An example of FPS: selecting 50 points from 1000 points in 2D space. red: currently selected point; green: previous selected points; blue: unselected points}
    \label{fig: FPS in 2D domain}
\end{figure}

\paragraph{FPS} The concept of Farthest Point Sampling (FPS) originates from the field of 3D computer vision \cite{qi2017pointnet}. Its primary goal is to select a diverse set of points from a given point cloud, which facilitates hierarchical feature extraction for downstream applications. The FPS process begins with a randomly selected initial point. In each subsequent iteration, a new point is chosen based on its maximum distance from all previously selected points as shown in Eq. \ref{eq:fps} where $S$ refers to the selected set, $R$ refers to the remaining set and $dist(i,j)$ represents the Euclidean distance between $i$ and $j$. For initialization, the selected set \( S \) is empty and the remaining set \( R \) equals to the entire dataset. An illustration of FPS is provided in Figure \ref{fig: FPS in 2D domain}, where 50 points are selected from a set of 1000 points in a 2D domain. Unlike some other selection methods, FPS solely relies on Euclidean distance to promote diversity among the selected points.

\begin{equation}
\label{eq:fps}
\operatorname*{argmax}_{i \in R} \left[ \max_{j \in S} \text{dist}(i, j) \right]
\end{equation}

\begin{figure}[t]
    \centering
    \includegraphics[width=0.6\textwidth]{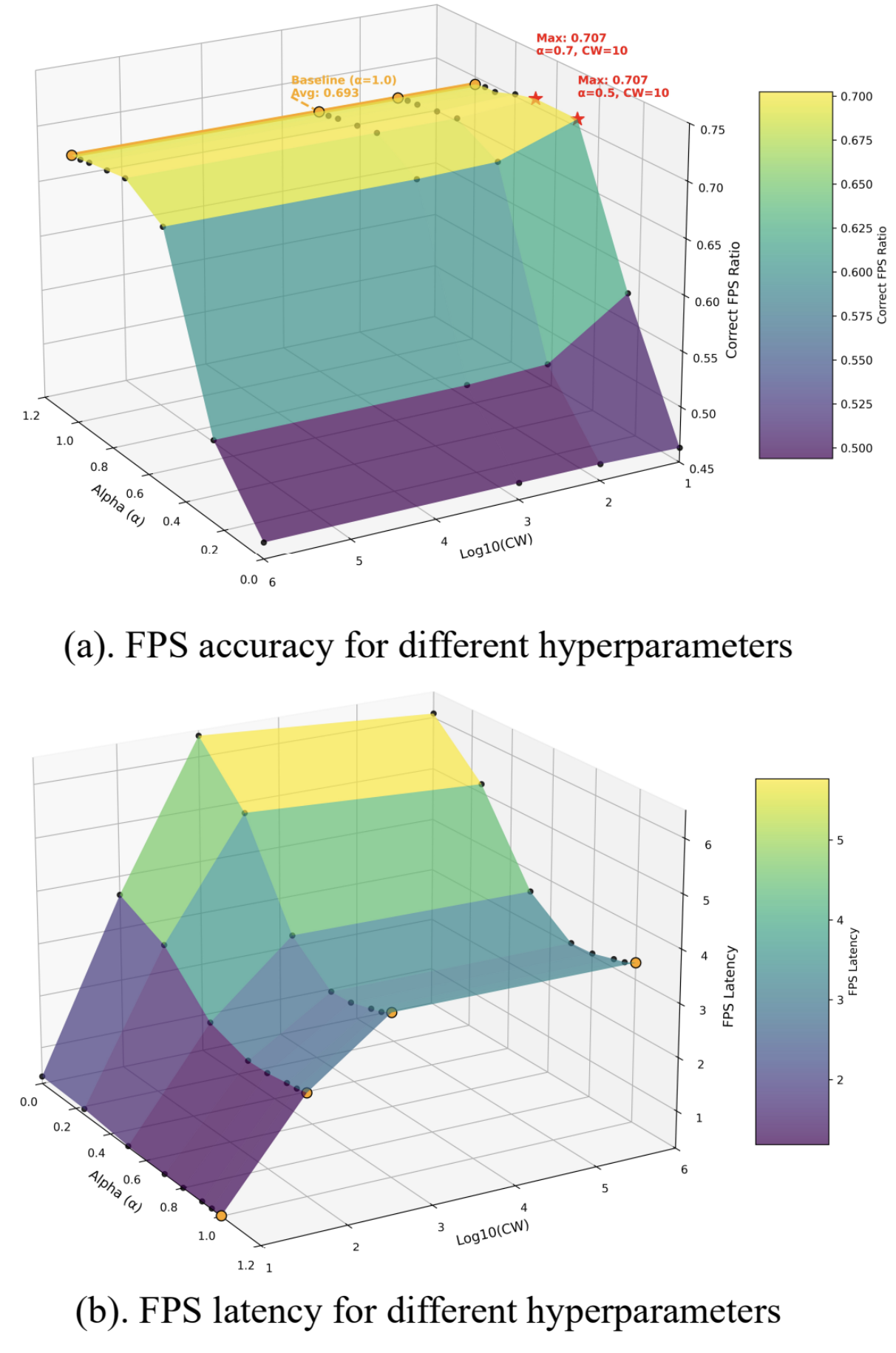}
    \caption{The impact of different hyperparameters: $\alpha$, $w$, $c_{r}$ on the recall accuracy and latency using FPS of the Natural Question dataset of single document Q\&A.}
    \label{fig: Natural_Question_hyperparameters_acc}
\end{figure}

\paragraph{MMR} In contrast, Maximal Marginal Relevance (MMR) selects a subset \( S \) from a large dataset with the dual objectives of maximizing rewards and ensuring diversity \cite{10.1145/290941.291025}. At each iteration, an element is chosen based on a locally optimal criterion defined in Eq. \ref{eq:mmr}. Here, the reward \( r_i \) of the \( i \)-th item is balanced against the similarity measure, where \( \text{cos}(i, j) \) denotes the cosine similarity between the \( i \)-th item and those in a context window \( W \) (comprising the most recently selected examples). This context window, for example, could include the last 10 selected samples when \( w = 10 \), reducing the influence of earlier selections. The iterative process continues until a predefined termination criterion, such as reaching a maximum number of tokens, is met.

\begin{equation}
\label{eq:mmr}
\operatorname*{argmax}_{i \in R} \left[ \alpha \cdot r_{i} - (1 - \alpha) \cdot \max_{j \in W} \text{cos}(i, j) \right]
\end{equation}

When comparing the two methods, both FPS and MMR use greedy selection to promote diversity. The key differences lie in their selection criteria: while FPS focuses exclusively on maximizing Euclidean distance for diversity, MMR introduces an additional reward factor and employs cosine similarity within a dynamic context window to balance between relevance and diversity. In this paper, we extend FPS by incorporating the reward mechanism and the dynamic context window. Ultimately, the key distinction between the modified FPS and MMR lies in their similarity metrics: the former uses Euclidean distance, while the latter relies on cosine similarity.

\subsection{Combine MMR and Modified FPS with LLM for Diversity on Q\&A and Summarization}
Extending MMR and Modified FPS techniques for LLMs in tasks such as Q\&A and summarization is relatively straightforward as shown in Fig. \ref{fig: placeholder}. These techniques employ a greedy approach to iteratively balance the similarity of selected sentences or chunks to the query with the diversity among the selected sentences or chunks. This method enhances the likelihood of selecting the most relevant sentences or chunks for LLMs in downstream tasks. Lastly, inspired by \cite{liu2023lostmiddlelanguagemodels}, a heuristic rearrangement scheme is implemented to enhance the likelihood of identifying the correct answer from the retrieved documents.

\paragraph{Q\&A} To evaluate the ability of LLMs on accurately extracting the correct answer, a query, a document, and a corresponding answer are initially provided. Documents are pre-processed by dividing them into sentences or chunks, and their embeddings are extracted beforehand. Both the query and the segmented documents are processed using encoder-only models to generate embeddings. In MMR, similarity is measured using the cosine angle, whereas in Modified FPS, Euclidean distance is used to assess similarity.  For benchmarking Q\&A performance, two metrics should be evaluated:
\begin{enumerate}
    \item Pre-LLM recall: whether the answer exists in the selected content before being sent to the LLM.
    \item Post-LLM recall: whether the answer appears in the LLM's output. 
\end{enumerate}
If the first metric shows significant improvement, the benefit of diversity becomes evident. Otherwise, the advantage of diversity may be limited. If the first metric improves while the second metric does not, it indicates that the performance of downstream tasks may be constrained by the capabilities of the LLM  \cite{liu2023lostmiddlelanguagemodels}.

\paragraph{Summarization}  
In summarization tasks, datasets typically consist of a document paired with a corresponding golden summary created by experts. When no specific query is provided, the process begins by dividing the document into manageable chunks. Encoder-only models are employed to generate embeddings for these chunks, and the mean of these embeddings is used to represent the query embedding. Following this, the same methodology as in the previous Q\&A task is applied to extract content that optimizes both reward and diversity.

The selected chunks are ordered to align with their original sequence in the document. These ordered chunks are sent to the LLM for summarization. We recognize that evaluating the extracted content before it is submitted to the LLM for summarization may not be particularly meaningful. Instead, we assess the quality of the LLM-generated summary by comparing it to the golden summary using metrics such as ROUGE \cite{lin-2004-rouge} or LLM-as-a-judge \cite{hsu2024rateexplainciterec}.

\begin{table*}
\centering
\small
\begin{tabular}{c|ccc|ccc|ccc}
\hline
\multirow{2}{*}{Q\&A} & \multicolumn{3}{c|}{Natural Question} & \multicolumn{3}{c|}{Trival Q\&A} & \multicolumn{3}{c}{Narrative Q\&A} \\
 & $c_r$=0.05 & $c_r$=0.1 & $c_r$=0.2 & $c_r$=0.05 & $c_r$=0.1 & $c_r$=0.2 & $c_r$=0.05 & $c_r$=0.1 & $c_r$=0.2 \\
\hline
SB & 46.28 & 58.60 & 69.41 & 63.44 & 71.64 & 78.33 & 18.60 & 21.04 & 25.61 \\
SB+MMR & 50.43 & 63.18 & \textbf{72.47} & \textbf{65.29} & \textbf{74.02}  & \textbf{80.47} & 20.88 & 24.09 & \textbf{27.59} \\
SB+FPS & \textbf{50.88} & \textbf{63.23} & 72.33 & 65.25 & 73.18 & 80.07 & \textbf{21.34} & \textbf{24.24} & 27.44 \\
\hline
\end{tabular}
\caption{This table compares the performance of SB+MMR and SB+FPS against SB across three different datasets and three compression ratios, focusing on the recall of the correct answer within the selected documents.}
\label{tab:Single_Document_Previous_LLM}
\end{table*}

\begin{table*}
\centering
\small
\begin{tabular}{c|ccc|ccc|ccc}
\hline
\multirow{2}{*}{Q\&A} & \multicolumn{3}{c|}{Natural Question} & \multicolumn{3}{c|}{Trival Q\&A} & \multicolumn{3}{c}{Narrative Q\&A} \\
 & $c_r$=0.05 & $c_r$=0.1 & $c_r$=0.2 & $c_r$=0.05 & $c_r$=0.1 & $c_r$=0.2 & $c_r$=0.05 & $c_r$=0.1 & $c_r$=0.2 \\
\hline
SB (index sort) & 40.44 & 51.74 & 64.25 & 75.66 & 75.95 & 76.25 & 15.40 & 17.98 & 18.60 \\
SB (sort) & 40.25 & 50.81 & 60.61 & 75.76 & 76.15 & 76.38 & 13.72 & 15.55 & 17.07 \\
SB (1:1) & 40.25 & 51.55 & 60.24 & 75.29 & 76.33 & 76.47 & 14.18 & 16.31 & 17.99 \\
SB (2:1) & 41.28 & 50.99 & 61.08 & 75.31 & 76.33 & 76.87 & 14.18 & 16.46 & 17.84 \\
SB (3:1) & 39.41 & 51.09 & 60.05 & 75.56 & 76.13 & 76.57 & 14.63 & 16.01 & 17.23 \\
\hline
SB+MMR (index sort) & 45.76 & 57.25 & \textbf{67.90} & 75.73 & 76.20 & 76.72 & \textbf{18.60} & \textbf{19.05} & \textbf{20.27} \\
SB+MMR (sort) & 44.45 & 55.19 & 64.35 & 76.20 & \textbf{76.85} & 77.00 & 16.31 & 16.46 & 17.38 \\
SB+MMR (1:1) & 45.48 & 55.57 & 63.60 & 76.23 & 76.77 & \textbf{77.34} & 16.46 & 16.62 & 16.92 \\
SB+MMR (2:1) & 45.20 & 55.29 & 63.32 & 75.90 & 76.20 & 76.65 & 15.09 & 15.55 & 16.01 \\
SB+MMR (3:1) & 43.80 & 55.85 & 63.51 & 76.05 & 76.72 & 76.92 & 16.16 & 16.92 & 16.31 \\
\hline
SB+FPS (index sort) & \textbf{46.79} & \textbf{59.12} & 67.71 & 75.93 & 76.13 & 76.60 & 17.68 & 17.68 & 19.05 \\
SB+FPS (sort) & 45.76 & 57.25 & 63.32 & 75.83 & 76.45 & 76.87 & 16.31 & 16.62 & 17.23 \\
SB+FPS (1:1) & 46.14 & 57.90 & 62.76 & 75.78 & 76.35 & 77.09 & 16.62 & 17.07 & 16.77 \\
SB+FPS (2:1) & 45.76 & 56.13 & 62.20 & 75.88 & 76.23 & 77.02 & 15.85 & 16.62 & 16.46 \\
SB+FPS (3:1) & 44.27 & 55.10 & 63.69 & \textbf{76.40} & 76.77 & 76.95 & 15.70 & 16.46 & 16.01 \\
\hline
\end{tabular}
\caption{This table compares the performance of SB+MMR and SB+FPS against SB across three different datasets and three compression ratios, focusing on the recall of the correct answer within the LLM responses.}
\label{tab:Single_Document_After_LLM}
\end{table*}

\section{Experiments}
The experiments conducted in this paper focus on three main topics: 
1. Single Document Question Answering (Q\&A), 
2. Multiple Documents Question Answering (Q\&A), and 
3. Single Document Summarization.

For Single Document Q\&A, the goal is to choose the correct answer from a set of candidate sentences within a single document. In Multiple Document Q\&A, all documents in the dataset are firstly divided into chunks and then combined, and a query is used to find the correct answers across the entire dataset. Because the dataset size is too large, approximation methods are used to enhance efficiency and speed. Specifically, two metrics are evaluated: 1. recall of the correct answer in the extracted document, and 2. recall of the correct answer in the LLM response. The benefit of diversity is primarily reflected in the improvement of the first metric, while performance improvements in Q\&A and summarization are mainly indicated by the second metric.

For summarization, various hyperparameters are considered in the optimization process:
\begin{enumerate}
    \item The weight balance between reward and diversity, denoted as $\alpha$,
    \item The context window size, $w$,
    \item The compression ratio, $c_{r}$, or the maximum number of selected tokens, $T_{\text{max}}$.
\end{enumerate}

\subsection{Single Document Q\&A}
For single document Q\&A, three datasets are included: 1. Natural Question \cite{kwiatkowski-etal-2019-natural}, 2. Trival QA \cite{joshi2017triviaqalargescaledistantly} and 3. Narrative QA \cite{kočiský2017narrativeqareadingcomprehensionchallenge}. For each dataset, it is composed of thousands of (query, document, answer) pairs where the answer exists within the document and answers the query. For each document, we split it into sentence using Spacy package \cite{Honnibal_spaCy_Industrial-strength_Natural_2020}. SentenceBERT (SB) is utilized as the encoder to extract embeddings from sentences in different experiments \cite{reimers2019sentencebert}. Then, different compression ratios, i.e., $c_{r}=0.05, 0.1, 0.2$ are utilized. Here, $c_{r}=0.05$ represents that the fraction of the number of selected tokens over the total number of tokens should be 0.05, i.e., the termination condition when 0.05 of all tokens are selected for answering the query. As for $\alpha$ and $w$, different hyperparameters are tested in a two-level iteration. The first coarse-level iteration utilizes $\alpha$ from [0, 0.25, 0.5, 0.7, 0.8, 0.9, 0.95, 1] and $w$ from [0, 10, 100, 1000, 1000000]. Then, the best performing from the first coarse-level iteration is selected. For the second granular-level iteration, it further divides the neighbors of the best performing first coarse-level hyperparameters and select the ones that have the best performance. An example of the best performing hyperparameters for FPS on recall accuracy and latency in the Natural Question dataset in the coarse level can be found in subfigure (a) of Fig. \ref{fig: Natural_Question_hyperparameters_acc}. More details on the impact of hyperparameters can be found in Fig. \ref{fig: acc_latency_all_hyperparameters}. In particular, for Natural question, $\alpha=0.55$ and $w=1$ is the best for MMR and $\alpha=0.5$ and $w=1$ is the best for FPS. For Narrative Q\&A, $\alpha=0.55$ and $w=5$ is the best for both MMR and FPS. For Trival Q\&A, $\alpha=0.6$ and $w=3$ is the best for MMR and $\alpha=0.7$ and $w=1$ is the best for FPS.

\begin{table*}[h!]
\small
\centering
\begin{tabular}{c|ccc|ccc}
\hline
\multirow{2}{*}{Natural Question} & \multicolumn{3}{c|}{GPT4} & \multicolumn{3}{c}{GPT3.5} \\
 & $T_{max}$=120k & $T_{max}$=50k & $T_{max}$=20k & $T_{max}$=10k & $T_{max}$=5k & $T_{max}$=2k \\
\hline
E5 & 70.7 & 69.4 & 68.5 & 66.2 & 64.2 & 57.4 \\
E5+MMR & \textbf{71.5} & \textbf{71.5} & \textbf{69.8} & \textbf{67.2} & \textbf{65.1} & \textbf{57.8} \\
\hline
\multirow{2}{*}{Narrative Q\&A} & \multicolumn{3}{c|}{GPT4} & \multicolumn{3}{c}{GPT3.5} \\
 & $T_{max}$=120k & $T_{max}$=50k & $T_{max}$=20k & $T_{max}$=10k & $T_{max}$=5k & $T_{max}$=2k \\
\hline
E5 & 13.42 & 10.06 & 6.7 & 4.88 & 4.88 & 4.57 \\
E5+MMR & \textbf{22.56} & \textbf{20.43} & \textbf{15.85} & \textbf{14.94} & \textbf{12.20} & \textbf{7.01} \\
\hline
\multirow{2}{*}{Trival Q\&A} & \multicolumn{3}{c|}{GPT4} & \multicolumn{3}{c}{GPT3.5} \\
 & $T_{max}$=120k & $T_{max}$=50k & $T_{max}$=20k & $T_{max}$=10k & $T_{max}$=5k & $T_{max}$=2k \\
\hline
E5 & 84.62 & 81.15 & 74.01 & 70.24 & 65.08 & 56.55 \\
E5+MMR & \textbf{88.99} & \textbf{85.81} & \textbf{82.24} & \textbf{78.57} & \textbf{74.01} & \textbf{65.08} \\
\hline
\end{tabular}
\caption{This table compares the performance of E5+MMR against E5 across three different datasets, focusing on the recall of the correct answer within the selected documents.}
\label{tab:Multiple_Document_Previous_LLM}
\end{table*}

\begin{table*}[h!]
\centering
\small
\begin{tabular}{c|ccc|ccc}
\hline
\multirow{2}{*}{Natural Question} & \multicolumn{3}{c|}{GPT4} & \multicolumn{3}{c}{GPT3.5} \\
 & $T_{max}$=120k & $T_{max}$=50k & $T_{max}$=20k & $T_{max}$=10k & $T_{max}$=5k & $T_{max}$=2k \\
\hline
E5 (index sort) & 45.7 & 50.5 & 52.7 & 45.9 & 49.2 & 45.4 \\
E5 (sort) & 55.2 & 56.6 & 54.6 & 51.5 & 50.5 & 47.8 \\
E5 (1:1) & 53.5 & 55.9 & 55.8 & 50.5 & 50.9 & 46.9 \\
E5 (2:1) & 54.5 & 57.5 & 56.4 & 51 & 50.5 & 47.1 \\
E5 (3:1) & 54.7 & 57 & 54.8 & 51.3 & 49.8 & 47.4 \\
\hline
E5+MMR (index sort) & 46.2 & 50.3 & 53.2 & 47.6 & 50.9 & 48.6 \\
E5+MMR (sort) & 56.4 & \textbf{57.9} & \textbf{57} & 51.3 & 51.4 & 47.7 \\
E5+MMR (1:1) & 55 & 57.2 & 55.9 & 50.8 & 50.7 & 47.8 \\
E5+MMR (2:1) & \textbf{57.2} & 55.3 & 56.2 & 50.7 & 52.2 & \textbf{48.7} \\
E5+MMR (3:1) & 56.3 & 56.4 & 55.6 & \textbf{51.6} & \textbf{52.3} & 47.4 \\
\hline
\end{tabular}
\caption{This table compares the performance of E5+MMR against E5 on Natural Question, focusing on the recall of the correct answer within the LLM responses.}
\label{tab:Multiple_Document_After_LLM_Natural_Question}
\end{table*}

\begin{table*}[h!]
\centering
\small
\begin{tabular}{c|ccc|ccc}
\hline
\multirow{2}{*}{Narrative Q\&A} & \multicolumn{3}{c|}{GPT4} & \multicolumn{3}{c}{GPT3.5} \\
 & $T_{max}$=120k & $T_{max}$=50k & $T_{max}$=20k & $T_{max}$=10k & $T_{max}$=5k & $T_{max}$=2k \\
\hline
E5 (index sort) & 10.37 & 9.15 & 8.54 & 5.18 & 4.57 & 4.88 \\
E5 (sort) & 10.67 & 10.59 & 8.23 & 6.1 & 4.57 & 5.18 \\
E5 (1:1) & 9.76 & 9.45 & 7.93 & 4.88 & 4.88 & 5.18 \\
E5 (2:1) & 10.06 & 9.15 & 7.93 & 5.18 & 3.96 & 5.18 \\
E5 (3:1) & 10.98 & 9.45 & 7.93 & 5.79 & 4.57 & 4.88 \\
\hline
E5+MMR (index sort) & 10.67 & 10.37 & \textbf{11.28} & 4.88 & 6.1 & 4.88 \\
E5+MMR (sort) & \textbf{12.8} & 10.67 & 10.37 & 6.1 & 6.1 & 4.27 \\
E5+MMR (1:1) & 11.89 & 10.06 & \textbf{11.28} & 6.4 & 6.71 & 4.88 \\
E5+MMR (2:1) & 11.59 & 10.67 & 10.98 & 6.4 & 5.79 & \textbf{5.49} \\
E5+MMR (3:1) & 11.89 & \textbf{10.98} & 10.06 & \textbf{6.7} & \textbf{7.01} & 4.88 \\
\hline
\end{tabular}
\caption{This table compares the performance of E5+MMR against E5 on Narrative Q\&A, focusing on the recall of the correct answer within the LLM responses.}
\label{tab:Multiple_Document_After_LLM_NarrativeQA}
\end{table*}

\begin{table*}[h!]
\centering
\small
\begin{tabular}{c|ccc|ccc}
\hline
\multirow{2}{*}{Trival Q\&A} & \multicolumn{3}{c|}{GPT4} & \multicolumn{3}{c}{GPT3.5} \\
 & $T_{max}$=120k & $T_{max}$=50k & $T_{max}$=20k & $T_{max}$=10k & $T_{max}$=5k & $T_{max}$=2k \\
\hline
E5 (index sort) & 74.21 & 73.21 & 73.12 & 65.18 & 64.19 & 64.68 \\
E5 (sort) & 73.81 & 73.91 & 72.42 & 63.59 & 65.08 & 65.57 \\
E5 (1:1) & 74.4 & 73.12 & 72.92 & 64.29 & 64.29 & 64.98 \\
E5 (2:1) & 73.81 & 73.81 & 72.72 & 64.09 & 64.58 & 64.29 \\
E5 (3:1) & 73.31 & 74.11 & 72.72 & 63.99 & 64.38 & 64.78 \\
\hline
E5+MMR (index sort) & 74.7 & 74.9 & \textbf{73.51} & 66.47 & \textbf{66.87} & 64.88 \\
E5+MMR (sort) & 74.9 & 74.8 & 73.12 & 64.29 & 65.57 & \textbf{66.07} \\
E5+MMR (1:1) & \textbf{75.2} & \textbf{75.5} & 73.31 & 65.38 & 65.38 & 65.08 \\
E5+MMR (2:1) & 74.6 & 74.11 & 72.62 & 64.88 & 65.77 & 65.38 \\
E5+MMR (3:1) & 74.7 & 74.31 & 73.02 & \textbf{65.67} & 65.48 & 64.98 \\
\hline
\end{tabular}
\caption{This table compares the performance of E5+MMR against E5 on Trival Q\&A, focusing on the recall of the correct answer within the LLM responses.}
\label{tab:Multiple_Document_After_LLM_TrivalQA}
\end{table*}

Based on the results across various datasets, we can assert that diversity significantly enhances the recall of the correct answer within the selected document, as demonstrated in Table \ref{tab:Single_Document_Previous_LLM}, showing an improvement of 2\% to 5\%. When the extracted sentences are summarized by GPT4 using the prompt shown in Figure \ref{fig: prompt for QA}, the advantages of SB+MMR and SB+FPS over SB alone remain evident, as shown in Table \ref{tab:Single_Document_After_LLM}. Additionally, we observe that the performance of Trivial Q\&A after LLM is better than the retrieved sentences, with a consistent result of approximately 76\%. This suggests that the performance is largely influenced by the LLM, possibly due to pretraining on Trivial Q\&A, even when the retrieved documents are provided. FPS, using distance as the evaluation metric, performs slightly worse than MMR, which uses cosine similarity. Moreover, MMR is faster than FPS because computing cosine similarity is quicker than Euclidean distance in Python, especially as the compression ratio increases, as shown in subfigure (b) of Figure \ref{fig: Natural_Question_hyperparameters_acc}. This conclusion generally holds true across different datasets. The speed advantage of MMR becomes more critical as the number of candidates increases with the dataset size. Consequently, MMR will be used in the multiple document comparison in the next section.

Inspired by the paper "Lost in the Middle" \cite{liu2023lostmiddlelanguagemodels}, we sorted the selected sentences by different methods. The term "index sort" refers to sorting the sentences in their original order within the document. In comparison, "SB (m:n)" refers to allocating the first selected m sentences with highest scores at the beginning, the next n sentences with highest scores at the end, and then another m sentences at the beginning, continuing this pattern until all sentences are allocated. Specifically, "SB (sort)" is equivalent to "SB (1:0)" and does not alter the sequence of selected sentences. As shown in Table \ref{tab:Single_Document_After_LLM}, SB (index sort) performs best because the original sequential information of the selected sentences in the document, despite missing some internal information, makes the most sense for GPT-4 in downstream tasks.

\subsection{Mutiple Documents Q\&A}
For multiple documents Q\&A, the same three datasets are utilized. In these datasets, the number of documents and the length of documents are relatively long, making it impractical to split each document into sentences. Instead, we follow the general framework of RAG to split each document into chunks of 512 tokens, with an overlapping ratio of 0.5 (i.e., 256 tokens) between any two adjacent chunks. To extract embeddings from these chunks and adhere to the standard pipeline of RAG, we apply the E5 model \cite{wang2024textembeddingsweaklysupervisedcontrastive}.
After applying the chunking strategy, the number of chunks can still reach nearly 1 million, which is impractical for exact search. To facilitate approximate search, principal component analysis (PCA) \cite{MACKIEWICZ1993303} is first applied to reduce the dimensionality of the embeddings, followed by clustering \cite{bishop2006pattern} to ensure the average number of chunks is less than 10k. Unlike single document Q\&A, we set the maximum number of tokens rather than the compression ratio as the threshold for the maximum number of tokens selected. Specifically, $T_{max}$ is set to 2k, 5k, or 10k for GPT-3.5 and 20k, 50k, or 120k for GPT-4. Other settings remain the same.
Different hyperparameters for $\alpha$ and $w$ are tested. For the Natural Questions dataset, $\alpha=0.9$ and $w=5$ yield the best results for GPT-3.5, while $\alpha=0.7$ and $w=5$ are optimal for GPT-4 in MMR. For Narrative Q\&A, $\alpha=0.8$ and $w=30$ are best for GPT-3.5, and $\alpha=0.7$ and $w=300$ are best for GPT-4 in MMR. For Trivia Q\&A, $\alpha=0.7$ and $w=20$ are best for GPT-3.5, and $\alpha=0.8$ and $w=300$ are best for GPT-4 in MMR.
From the results, we observe that the optimal values for $\alpha$ and $w$ are generally larger for multiple document Q\&A compared to single document Q\&A.

When evaluating the performance of multiple-document Q\&A systems, we observe a pattern similar to that of single-document Q\&A. Specifically, the E5+MMR method shows a significant improvement over E5 in recall of the answers in retrieved documents, as demonstrated in Table \ref{tab:Multiple_Document_Previous_LLM}, with a margin exceeding 10\%. Additionally, E5+MMR outperforms E5 for post-LLM recall as shown in Tables \ref{tab:Multiple_Document_After_LLM_Natural_Question}, \ref{tab:Multiple_Document_After_LLM_NarrativeQA}, and \ref{tab:Multiple_Document_After_LLM_TrivalQA}. However, future research should prioritize enhancing the LLM's ability to utilize the retrieved documents effectively, rather than merely focusing on retrieving more accurate documents, as the LLM itself is the bottleneck.
This observation is further corroborated in Trivial Q\&A, where the results consistently achieve 64\% accuracy for GPT3.5 and 76\% for GPT4, irrespective of the retrieved document.
Last, unlike single-document Q\&A, placing important chunks at the beginning and ending positions of the prompt can provide benefits, particularly in Natural Question scenarios, as shown in Table \ref{tab:Multiple_Document_After_LLM_Natural_Question}, which can lead to a 10\% improvement. This finding aligns with the conclusions of the paper "Lost in the Middle". The most relevant chunks to the query should be positioned either at the beginning or the end of the prompt.

\begin{table*}[h!]
\centering
\small
\begin{tabular}{c|ccc|ccc|ccc}
\hline
\multirow{2}{*}{SquAD} & \multicolumn{3}{c|}{Sentence} & \multicolumn{3}{c|}{Chunk size: 256} & \multicolumn{3}{c}{Chunk size: 512} \\
 & 10k & 5k & 2k & 10k & 5k & 2k & 10k & 5k & 2k \\
\hline
SB/E5 & 86.8 & 83.7 & 78.5 & 95 & 92.7 & 86.3 & 96.7 & 94.3 & 86.6 \\
SB/E5+MMR & \textbf{90.1} & \textbf{89.4} & \textbf{86.6} & \textbf{97} & \textbf{96.6} & \textbf{95.4} & \textbf{99} & \textbf{97.8} & \textbf{96.7} \\
\hline
\end{tabular}
\caption{This table compares the performances of sentence splitter and chunk splitter of size 256 and 512 on SquAD, focusing on the recall of the correct answer within the selected documents.}
\label{tab:Squad_Document_Previous_LLM}
\end{table*}

\begin{table*}[h!]
\centering
\small
\begin{tabular}{c|ccc|ccc|ccc}
\hline
\multirow{2}{*}{SquAD} & \multicolumn{3}{c|}{Sentence} & \multicolumn{3}{c|}{Chunk size: 256} & \multicolumn{3}{c}{Chunk size: 512} \\
 & 10k & 5k & 2k & 10k & 5k & 2k & 10k & 5k & 2k \\
\hline
SB/E5 (index sort) & 70.4 & 73 & 71.7 & 78.2 & 81.1 & 79.2 & 80.2 & 82.9 & 78.2 \\
SB/E5 (sort) & 71.3 & 70.1 & 67.9 & 79.9 & 80 & 76.6 & 82.9 & 82.2 & 76.5 \\
SB/E5 (1:1) & 72 & 71.8 & 69.5 & 80.3 & 80.4 & 77.5 & 82.7 & 82 & 77.4 \\
SB/E5 (2:1) & 71.9 & 71.7 & 69.4 & 80.4 & 80.1 & 77.4 & 82.9 & 81.9 & 77.5 \\
SB/E5 (3:1) & 71.8 & 71 & 68.7 & 81.4 & 79.9 & 77.7 & 82.7 & 82.2 & 77.3 \\
\hline
SB/E5+MMR (index sort) & 68.7 & \textbf{74.2} & \textbf{75} & 76.5 & 81.4 & 83 & 76.1 & 84.2 & 84.9 \\
SB/E5+MMR (sort) & 71.5 & 71.8 & 72.5 & 82 & 81.9 & 82.6 & 84.1 & \textbf{84.5} & 84.7 \\
SB/E5+MMR (1:1) & \textbf{72.5} & 73.6 & 72.9 & 81.3 & 82.2 & 81.7 & 82.3 & 82.9 & 84.6 \\
SB/E5+MMR (2:1) & 71.9 & 73.1 & 73.2 & \textbf{81.8} & \textbf{82.5} & \textbf{83.1} & \textbf{84.5} & \textbf{84.5} & \textbf{85} \\
SB/E5+MMR (3:1) & 71.7 & 73.4 & 71.9 & 81.4 & \textbf{82.5} & 82.9 & 83.4 & 84.3 & 84.5 \\
\hline
\end{tabular}
\caption{This table compares the performances of sentence splitter and chunk splitter of size 256 and 512 on SquAD, focusing on the recall of the correct answer within the LLM responses.}
\label{tab:Squad_Document_After_LLM}
\end{table*}

\begin{table*}[h!]
\centering
\small
\begin{tabular}{c|cc|cc}
\hline
\multirow{2}{*}{Summarization Datasets} & \multicolumn{2}{c|}{gov\_report} & \multicolumn{2}{c}{legal} \\
 & Rouge & GPT4 WR & Rouge & GPT4 WR \\
\hline
SB & 17.7 & 24.24 & 11.3 & 35.82 \\
SB+MMR & \textbf{18} & $\frac{72.65+77.86}{2}=\textbf{75.26}$ & \textbf{11.8} & $\frac{71.64+56.72}{2}=\textbf{64.18}$ \\
\hline
\end{tabular}
\caption{This table compares the performance of SB+MMR against SB on gov\_report and legal, using ROUGE and LLM-as-a-judge.}
\label{tab:Summarization}
\end{table*}

\subsection{Sentence and Chunk Splitter Comparison on SquAD}
For the comparison between sentence and chunk splitters on multiple documents Q\&A, only the SQuAD dataset will be considered. The dataset sizes for Natural Questions, TriviaQA, and NarrativeQA are too large, making sentence-level experiments difficult.
For the sentence-level splitter, Spacy is used. For the chunk-level splitter, a threshold of 256 or 512 tokens with 50\% overlap between adjacent chunks is applied. All segmented sentences or chunks are mixed, reduced in dimension through PCA, and clustered for downstream tasks. Similar to previous experiments, $T_{max}$ is set to 2k, 5k, or 10k for GPT3.5. For the sentence-level splitter, the best parameters are $\alpha=0.25$ and $w=1000$. For the 256-token chunk-level splitter, the best parameters are $\alpha=0.5$ and $w=300$. For the 512-token chunk-level splitter, the best parameters are $\alpha=0.3$ and $w=300$.
The results are consistent with previous findings. SB/E5+MMR significantly outperforms SB/E5, as shown in Table \ref{tab:Squad_Document_Previous_LLM}, with a 10\% increase in recall of the correct answer within the selected documents. This recall increment of SB/E5+MMR over SB/E5 still exists in the LLM response, as shown in Table \ref{tab:Squad_Document_After_LLM}. "Index sort" generally performs better for sentence-level splitting, while sorting based on score is usually beneficial for chunking. A new takeaway is that chunk-level performance is better than sentence-level, with even better results for larger chunk sizes.

\subsection{Single Document Summarization}
For single document summarization, we include two datasets: the gov report \cite{huang2021efficientattentionslongdocument} and legal documents \cite{shukla2022legalcasedocumentsummarization}. We utilize GPT3.5 for summarization. To achieve this, we filter examples that are less than 15k tokens and then apply MMR to select sentences within each document until it reaches the predetermined threshold of 8k tokens. For the 1. gov report, the best parameters are 
$\alpha=0.9$ and $w=10$. For the legal documents, the best parameters are $\alpha=0.925$ and $w=300$.
After selecting and ordering the selected sentences based on their original sequence, they are sent to GPT3.5 to generate the final summary using a specific prompt in Figure. \ref{fig: prompt for Summarization}. For both datasets, expert-written golden summaries are provided for each document. We evaluate the quality of generated summary using the ROUGE score by comparing with the golden summary. In addition, summaries by SB and SB+MMR are compared using LLM-as-a-Judge through GPT4. To address the position bias problem, we switch the sequences of the two summaries in two runs and average the win rate (WR). Our experiments reveal that diversity improves summary quality, as indicated by increased ROUGE scores and a higher LLM-as-a-Judge WR.
Additionally, experiments on our internal data show that diversity is particularly beneficial for long emails, articles, and logs, where redundancy is a significant issue due to repetitive content, greetings, and long URLs. Diversity avoid overestimating information similar to the query.

\section{Related Works}

\subsection{Surpass the Context Window Limitation}
The quadratic time complexity \(O(N^{2})\) of self-attention in transformer \cite{vaswani2023attention} poses limitations on training larger context window \cite{ouyang2022training,openai2024gpt4}. A wide range of approaches has been explored to extend the context window of these models. One approach, exemplified by Longformer and its subsequent iterations \cite{beltagy2020longformer}, aims to handle much longer contexts by modifying the self-attention mechanism. Another strategy involves training models on shorter sequences and then modifying the Transformer's structure to accommodate longer sequences during inference \cite{press2022train}. Lastly, some researchers are exploring methods such as chunking, leveraging discourse structures, and extracting key content to effectively reduce the context window required by LLMs, thereby facilitating the summarization of extensive documents \cite{Koh_2022}. Our paper will focus on the extraction method.

\subsection{RAG \& Long Document Summary}
To give LLM the capability of utilizing unknown data, RAG is developed to retrieve relevant materials and utilize them for downstream tasks \cite{lewis2021retrievalaugmentedgenerationknowledgeintensivenlp}. Several downstream approaches enhance RAG from various perspectives including graphRAG \cite{edge2025localglobalgraphrag}, correctiveRAG \cite{yan2024correctiveretrievalaugmentedgeneration}, hierarchical RAG \cite{sarthi2024raptorrecursiveabstractiveprocessing}, modular RAG \cite{gao2024modularragtransformingrag} and more \cite{10.2308/ISYS-2023-047}.

Long document summary includes extractive based method and abstractive based method. Initially, extractive methods, such as unsupervised Tf-Idf \cite{liang-etal-2021-improving} and supervised BERT \cite{devlin2019bert}, were widely adopted until the advent of more potent decoder-based models. Subsequently, abstractive approaches gained prominence, exemplified by the end-to-end method \cite{lewis2019bart} and hybrid extractive-abstractive strategies\cite{zhang2020pegasus, Yuan_Wang_Cao_Li_2023, xie2022gretel, lim-song-2023-improving}.

Recently, Hypothetical Document Embedding (HyDE) applied LLM reranking to enhance diversity in Q\&A tasks, claiming their method outperforms MMR, which provides nearly no benefit on downstream tasks \cite{eibich2024aragogadvancedragoutput, pickett2024betterragusingrelevant}. However, utilizing LLM for reranking can be expensive, slow and costly. In addition, these studies did not discuss the recall of relevant documents prior to LLM generation and explore various hyperparameters within MMR. This motivates us to explore more on MMR and FPS to enhance diversity, thereby improving the performance of LLMs on downstream tasks.

\subsection{Representative Sentences Extraction}
In the realm of extraction-based summarization, the criterion for choosing key sentences is crucial. Tf-Idf selected sentences based on the frequency of keyword \cite{liang-etal-2021-improving}. PEGASUS selected sentences that utilizes structural information and ROUGE scores \cite{zhang2020pegasus}. BERT selected representative sentences by distance in embedding space \cite{devlin2019bert}. SentenceBERT \cite{reimers2019sentencebert}, DeBERTa \cite{he2021debertadecodingenhancedbertdisentangled}, E5 \cite{wang2024textembeddingsweaklysupervisedcontrastive} and DPR \cite{karpukhin2020densepassageretrievalopendomain} have been trained on larger corpora using more powerful computational resources. What is more, these methods only focus on selecting sentences that are related with core idea of the document, and it will cause problem for lacking the diversity of selected sentences. This research primarily focuses on leveraging smaller encoder-only models while incorporating diversity as a key consideration.

\section{Conclusion}
This study proves the benefits of diversity through MMR and FPS to LLM performances on Q\&A and summarization. From the retrival viewpoint, the recall is greatly improved both for sentence and chunk-level splitter, especially when $\alpha$ and $w$ are properly selected. This recall rate increment is maintained after LLM generation. However, future research should pay more attention to improve the LLM's capability to find answers from the retrieved documents. MMR shows slightly better performances compared with FPS, and its latency property is much better, which greatly increase the potential of usage in application. For sentence-level splitter, arranging the selected sentences in their original sequence is usually beneficial and for chunk-level splitter, putting more important chunks at the beginning and ending positions are beneficial. Lastly, given a multiple document Q\&A like SquAD, chunk-level splitter usually has a better performance compared with sentence-level splitter. Lastly, these conclusion on Q\&A can be extended to summarization task. 

\section{Limitation}
There are several limitation on this works. To begin with, we only work on English dataset, while multilingual datasets should be tested to prove the importance of diversity on other language. In addition, this work focuses on research dataset while more work is supposed to be conducted on industrial datasets. Next, for extremely large dataset, more engineering work on parallelization like tree structures should be conducted to reduce latency. Finally, further research is needed to enhance LLMs' ability to effectively utilize retrieved documents—such as through post-training methods that encourage the model to place greater emphasis on the retrieved information.

\bibliographystyle{unsrt}  
\bibliography{references}  

\newpage
\appendix

\section{Prompts for Q\&A and Summarization}

\begin{figure*}[h!]
    \centering
    \includegraphics[width=0.9 \textwidth,height=\textheight,keepaspectratio]{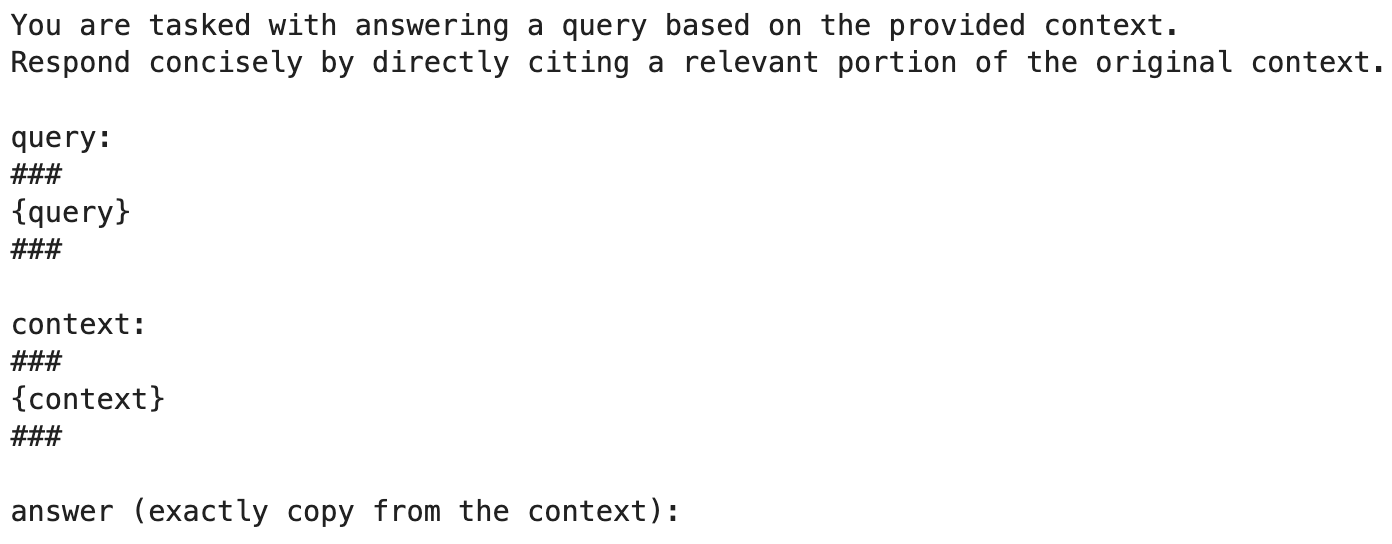}
    \caption{Prompts for Q\&A}
    \label{fig: prompt for QA}
\end{figure*}

\begin{figure*}[h!]
    \centering
    \includegraphics[width=0.9 \textwidth,height=\textheight,keepaspectratio]{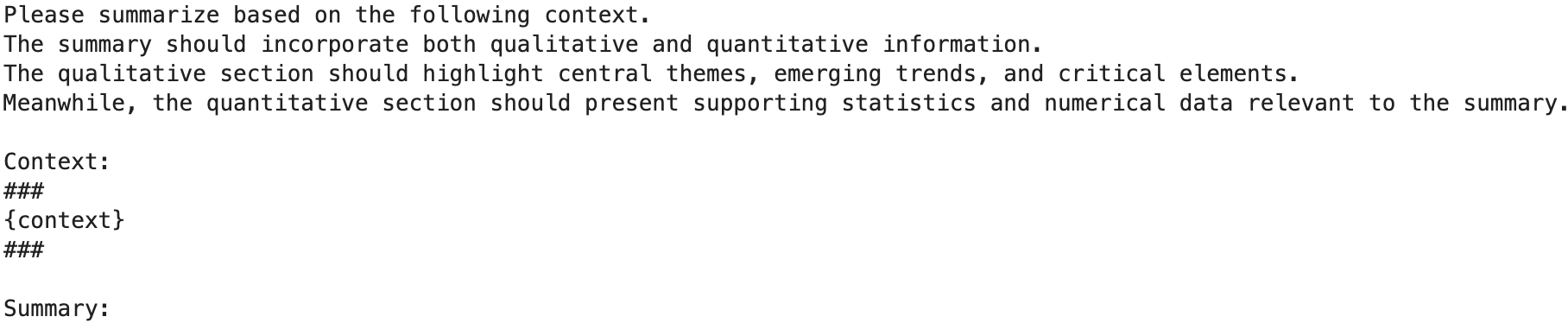}
    \caption{Prompts for Summarization}
    \label{fig: prompt for Summarization}
\end{figure*}

Here are the prompts for Q\&A in Figure. \ref{fig: prompt for QA} and summarization in Figure. \ref{fig: prompt for Summarization}.

\section{Impacts of Hyperparameters in Diversity for Retrieval}
\begin{figure*}[h!]
    \centering
    \includegraphics[width=\textwidth,height=\textheight,keepaspectratio]{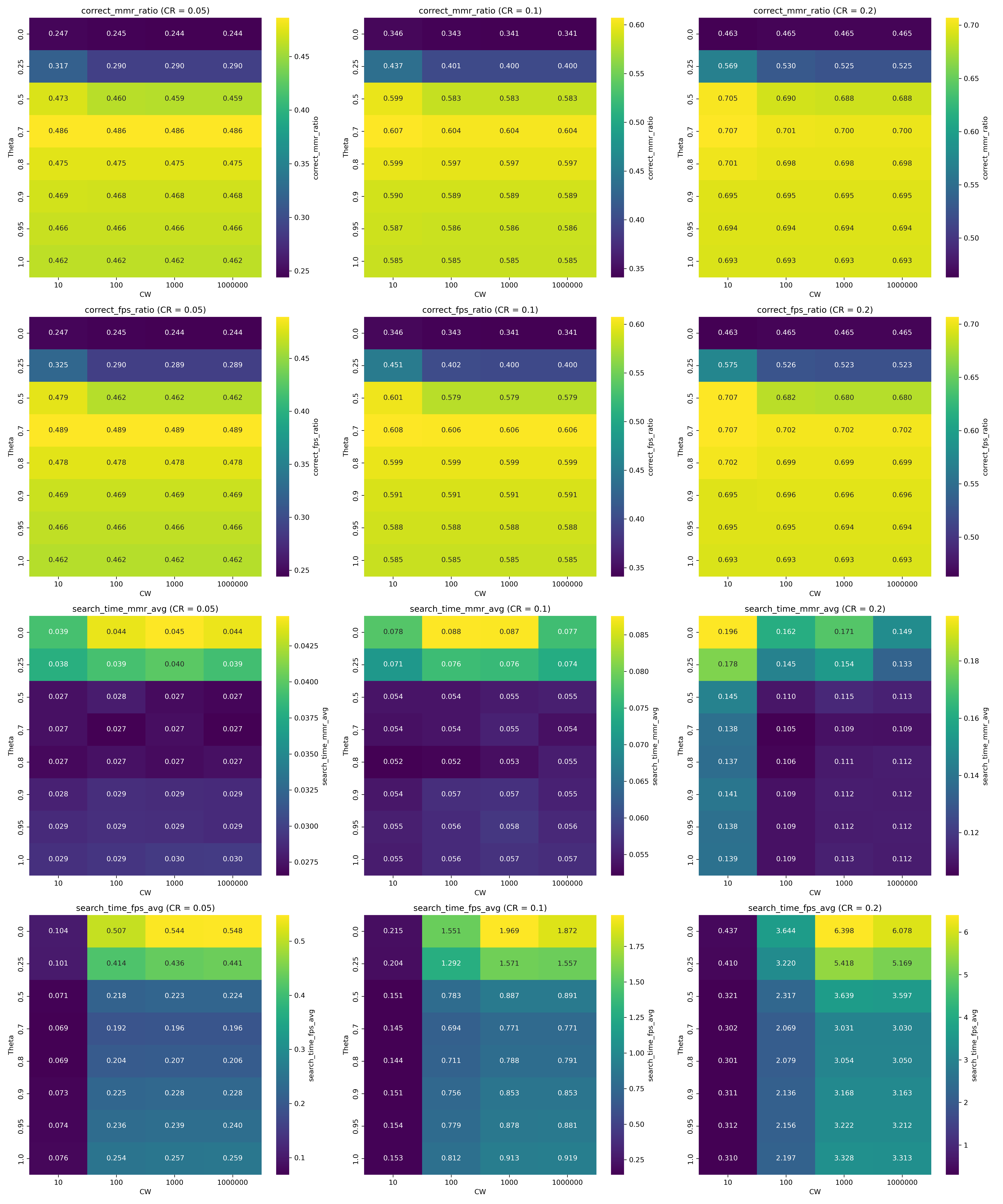}
    \caption{The impact of different hyperparameters: $\alpha$, $w$, $c_{r}$ on the recall of the Natural Question dataset of single document Q\&A.}
    \label{fig: acc_latency_all_hyperparameters}
\end{figure*}

The impact of different hyperparameters: $\alpha$, $w$, $c_{r}$ on the recall of the Natural Question dataset of single document Q\&A is shown in Figure. \ref{fig: acc_latency_all_hyperparameters}. The first and second subfigures illustrate the recall ratios of answers contained in the selected documents for SB+MMR and SB+FPS. When the weight parameter \( w = 1 \), they are equivalent to SB. From the results, we can conclude that both SB+MMR and SB+FPS outperform SB. The last two subfigures display the latency of SB+MMR and SB+FPS. SB+FPS shows slightly worse performances than SB+MMR, and the latency of SB+MMR is significantly lower, especially when the context window is very long. Considering these two aspects, SB+MMR is more suitable for practical use compared to SB+FPS.

\end{document}